\documentclass[runningheads]{llncs}
\usepackage{booktabs}
\usepackage{amssymb}
\usepackage[T1]{fontenc}
\usepackage{graphicx}
\usepackage{url}
\newcommand{\squishlist}{\begin{itemize}}
\newcommand{\squishend}{\end{itemize}  }

\newcommand{\squishenum}{
\begin{list}{$\bullet$}{ 
    \setlength{\itemsep}{1pt}
    \setlength{\parsep}{0pt}
    \setlength{\topsep}{1.5pt}
    \setlength{\partopsep}{0pt}
    \setlength{\leftmargin}{2em}
    \setlength{\labelwidth}{1.5em}
    \setlength{\labelsep}{0.5em} } }

\usepackage{verbatim}

\usepackage{xcolor}

\newcommand{\ben}[1]{\textcolor{black}{#1}}

\usepackage{multirow}
\usepackage{adjustbox} 
\usepackage{arydshln}

\usepackage{relsize}
\usepackage{subcaption}

\usepackage{graphicx}
\usepackage{amsmath}

\usepackage{todonotes}
\usepackage{xcolor}
\usepackage{ifthen}

\newboolean{includeMemo}
\setboolean{includeMemo}{true} 

\newcommand{\memo}[1]{\ifthenelse{\boolean{includeMemo}}{\todo[inline,caption={},color=green!20!]{#1}}}
\newcommand{\memob}[1]{\ifthenelse{\boolean{includeMemo}}{\todo[inline,caption={},color=blue!20!]{#1}}}
\usepackage{tikz}
\usetikzlibrary{positioning,calc,decorations.pathreplacing,arrows.meta}

\begin{document}
\title{Symbolic Guidance for LLM Agents in Distributed Multiagent Coordination\thanks{A preliminary version of this work was published as an extended abstract at AAMAS 2026~\cite{rachmut2026symbolic}.}}
%
%
\author{Ben Rachmut \and
Ning Zhang \and
Yevgeniy Vorobeychik \and
William Yeoh}

\authorrunning{B. Rachmut et al.}

\institute{Washington University in St. Louis, Saint Louis, MO, United States\\
\email{\{benr, zhang.ning, yvorobeychik, wyeoh\}@wustl.edu}}
%
%
%
\maketitle              
\sloppy
\allowdisplaybreaks

\begin{abstract}
Large language models (LLMs) are increasingly deployed as autonomous agents within multi-agent systems. Yet, their capacity to execute distributed coordination protocols remains poorly understood. Recently, researchers introduced AgentsNet, a framework that enables LLM agents to coordinate in a distributed manner to solve a range of multi-agent coordination problems. However, because this approach grants agents full flexibility in reasoning about their coordination strategies, it often yields inconsistent or poor performance, particularly in more complex domains.

In this paper, we hypothesize that LLM agents can achieve better coordination when provided with symbolic guidance derived from established symbolic algorithms. To test this hypothesis, we systematically evaluate how different forms of guidance influence performance across three canonical distributed graph-based problems -- graph coloring, matching, and vertex cover. For each problem, we examine two variants: A simpler one where feasible solutions can be found using local, agent-based heuristics, and a more complex one where optimal solutions require global coordination.

To structure this investigation, we introduce the \emph{Symbolic Guidance Taxonomy (SGT)}, which defines a spectrum of guidance ranging from the AgentsNet baseline -- using only freeform natural-language task descriptions with no explicit guidance -- at one end, to complete algorithmic specifications at the other, with intermediate levels incorporating partial pseudocode. Our results reveal that intermediate guidance levels are most effective: Partial pseudocode guidance consistently outperforms the unguided AgentsNet baseline. Moreover, both the choice of model and source of symbolic guidance play important roles. Gemini agents generally achieve the best overall performance; Llama agents often benefit from guidance; whereas Qwen agents tend to be largely insensitive to it. Similarly, guidance derived from local heuristic algorithms proves broadly robust, while guidance from complex global search algorithms tends to be less effective. Collectively, these findings offer design principles for balancing symbolic structure and LLM adaptability in distributed multi-agent coordination.

\keywords{Large Language Models \and Multiagent Systems \and Distributed Coordination \and Symbolic Guidance \and Adjustable Autonomy}

\end{abstract}

\section{Introduction}

Coordinating multiple autonomous agents to achieve a common objective is a long-standing and fundamental challenge in AI. Whether in scheduling meetings~\cite{maheswaran:04a}, deploying multi-robot teams~\cite{pertzovskiy2023cams}, managing smart homes and smart grids~\cite{miller:12,fioretto:17a,rust2022resilient}, or controlling satellite constellations~\cite{ZilbersteinRSC24}, the core difficulty lies in ensuring that coherent global behavior emerges from local decisions made under limited information and communication. Such multiagent coordination problems are inherently complex and computationally demanding, requiring agents to reason about both their own actions and those of others in dynamic, distributed environments.

Over the past several decades, symbolic approaches have provided elegant solutions to these problems. Distributed algorithms, many of which were proposed by the AAMAS community~\cite{NguyenYLZ19,Barde2024ModelBasedOfflineCoordination,modi2005adopt,cohen2017max}, have enabled agents to coordinate through rigorously defined message-passing and optimization procedures, often with formal guarantees of correctness and convergence. However, these symbolic solvers depend on complete and well-specified problem formulations, making them fragile when the environment or objectives are uncertain, underspecified, or expressed informally.

In parallel, the recent emergence of agentic AI systems, that is, large language models (LLMs) acting as autonomous agents, has opened a new frontier for multiagent coordination \cite{park2023generative,wang2024survey,groetschla2025agentsnet}. LLM agents can interpret instructions and interact in natural language, making them robust to incomplete, ambiguous, or imprecise specifications. Yet, when deployed off the shelf, they struggle to coordinate reliably, particularly in domains where global coherence must arise from local interactions \cite{groetschla2025agentsnet}. In contrast to symbolic algorithms, LLM agents lack built-in mechanisms to ensure consistency, convergence, or optimality.

This contrast exposes a core tradeoff between the two paradigms. Symbolic solvers offer theoretical rigor and reliability but limited flexibility, whereas LLM agents provide linguistic adaptability and robustness but lack structure for consistent coordination. Bridging this gap requires combining the complementary strengths of both.

To this end, we propose integrating symbolic guidance into LLM-based multiagent systems and systematically studying its effects. Specifically, we introduce the \emph{Symbolic Guidance Taxonomy (SGT)}: A principled framework that characterizes different levels and forms of symbolic scaffolding for LLM agents. Inspired by ideas from adjustable autonomy \cite{scerri2002towards,scerri2001adjustable}, SGT positions LLM agents along an autonomy spectrum: From fully open-ended reasoning (maximal autonomy), to strict algorithmic adherence (minimal autonomy). 
This taxonomy provides a unified lens for understanding how symbolic structure can guide agentic reasoning, revealing when and how symbolic scaffolding enhances coordination performance.

Building on this analogy, we use SGT not only as a conceptual lens but also as the basis for our evaluation methodology. Specifically, we instantiate benchmark graph-based coordination problems -- graph coloring, matching, and vertex cover -- each encompassing both local and global objectives. Local objectives (e.g.,~coloring with $\Delta + 1$ colors, where $\Delta$ is the graph’s maximum degree; maximal matching; minimal vertex cover) capture feasible but suboptimal coordination under partial information, whereas global objectives (e.g.,~3-coloring, maximum matching, minimum vertex cover) impose stronger optimality requirements. Our experimental setup systematically factors both algorithmic structure (e.g.,~local agent-based heuristics versus global distributed search procedures) and symbolic guidance levels as defined by SGT, enabling direct comparison across LLM families, task types, and autonomy regimes.

Our results show that symbolic guidance profoundly shapes coordination outcomes: Structured scaffolding significantly improves coherence and stability on harder global objectives, while excessive prescriptiveness can reduce adaptability in simpler tasks. These findings highlight that the balance between symbolic control and LLM autonomy is neither binary nor monotonic, but context-dependent, mirroring the broader principle of adjustable autonomy in human-agent collaboration.

In summary, our key contributions are as follows:

\begin{enumerate}
\item \textbf{The Symbolic Guidance Taxonomy (SGT):} A principled framework for structuring guidance strategies in multi-agent LLMs, inspired by adjustable autonomy and formalized in terms of algorithmic selections and representational encodings.
\item \textbf{A unified evaluation framework} that systematically combines benchmark distributed graph problems, controllable difficulty levels (local vs. global objectives), and structured variation in symbolic guidance.
\item \textbf{An analysis of algorithm-guidance interactions} that disentangles when LLMs benefit from symbolic structure versus when open-ended reasoning yields superior adaptability.
\item \textbf{Empirical design principles} for balancing symbolic control and LLM flexibility, providing actionable insights for deploying multiagent LLM systems in real-world distributed environments.
\end{enumerate}

\section{Background} \label{sec:background}

Distributed coordination problems are naturally formalized as graphs, where each agent corresponds to a vertex and communicates through message exchange with its neighboring vertices. The central challenge lies in ensuring that these locally informed decisions collectively produce a globally consistent outcome. To systematically examine this challenge, we first describe the three canonical benchmark domains that is the focus of this paper: \emph{Graph coloring}, \emph{matching}, and \emph{vertex cover}. We then contrast two classical algorithmic paradigms: \emph{Local agent-based heuristics}, which emphasize scalability and decentralized decision-making, and \emph{global distributed search procedures}, which provide stronger guarantees of optimality but demand greater coordination and communication overhead. We describe a representative algorithm from each of the two paradigms that we use in this paper. 

\subsection{Benchmark Domains and Variants} 
\label{sec:domains}

We now describe the three canonical benchmark domains that we use in this paper: Graph coloring, matching, and vertex cover. For each of these domains, we describe a \emph{local variant}, where neighborhood information is sufficient to solve the problem, and a \emph{global variant}, where consistency must be enforced across the entire network~\cite{peleg2000distributed}.

\paragraph{\textbf{Graph Coloring:}} In graph coloring, each agent must choose a color distinct from its neighbors. If the palette has $(\Delta+1)$ colors, where $\Delta$ is the maximum degree, every agent can resolve conflicts locally~\cite{linial1992locality}. This is the \emph{local} variant. But with only three colors, local choices no longer suffice: Agents must coordinate \emph{globally} to avoid contradictions~\cite{blum1994new}. 

\paragraph{\textbf{Matching:}} In matching, each agent must decide whether to form a pair with one of its neighbors. Stopping once no more pairs can be added yields a \emph{maximal matching}, which can be obtained through only \emph{local} agreements~\cite{hanckowiak2001distributed}. By contrast, finding a \emph{maximum matching}, that is, pairing as many agents as possible, requires \emph{global} coordination across the whole network~\cite{bar2017distributed}. 

\paragraph{\textbf{Vertex Cover:}} In vertex cover, agents must decide who takes responsibility for covering their incident edges. A cover is \emph{minimal} if every selected agent is essential, that is, removing any one of them would leave some edge uncovered~\cite{linial1992locality}. This corresponds to the \emph{local} variant. In contrast, finding a \emph{minimum} cover, where the number of selected agents is as small as possible, requires balancing responsibilities \emph{globally} across the network~\cite{parnas2007approximating}.

\subsection{Algorithmic Approaches}
\label{sec:algorithms}

We now describe a representative algorithm from the two contrasting algorithmic paradigms for distributed problems: \emph{Local agent-based heuristics} and \emph{global distributed search procedures}. 

\paragraph{\textbf{Local Agent-based Heuristics:}} We use the \emph{Maximal Independent Set (MIS)} heuristic, a cornerstone of distributed graph algorithms~\cite{luby1985simple,alon1986fast,linial1992locality,johansson1999simple}, as a representative approach from this category. An MIS is a set of vertices with no two adjacent, such that no additional vertex can be added without breaking independence. It can be computed using only local neighborhood information: Each vertex joins the set if none of its higher-priority neighbors has already done so. This simple ordering principle extends naturally across our three domains: In graph coloring, vertices commit to colors once higher-priority neighbors have chosen; in matching, two vertices form a pair only if neither has already committed to another neighbor; and in vertex cover, priorities resolve redundancy so that higher-priority vertices remain in the cover, ensuring every edge is covered.  These MIS-based rules admit efficient distributed algorithms running in $O(\log n)$ rounds, where $n$ is the number of nodes in the graph~\cite{linial1992locality,alon1986fast}.


\paragraph{\textbf{Global Distributed Search Procedures:}} We use \emph{Distributed Depth-First Search (DFS)} as a representative algorithm from this category. Distributed DFS enforces a strict global order: Agents are considered sequentially, and tentative choices are undone whenever conflicts arise. This guarantees optimality: Colorings are extended until a consistent assignment is found, matchings are branched on until the maximum is recovered, and vertex covers are explored until a minimum set is identified.  The same approach is also referred to as \emph{backtracking search} in the literature on distributed constraint satisfaction problems, where agents cooperatively assign values to variables subject to global satisfaction constraints~\cite{yokoo2002distributed}. 
However, such exhaustive search requires a runtime of $O(b^n)$, where $b$ is the branching factor for each agent (e.g., number of possible colors in graph coloring) and $n$ is the number of agents in the problem. Due to the exponential complexity, such approaches may be impractical for large-scale distributed coordination.

\section{Related Work}

Researchers have recently explore whether LLMs can coordinate effectively when deployed as interacting agents. \emph{AgentsNet} introduced the first systematic benchmark for LLM-based multi-agent coordination, evaluating problems such as $(\Delta+1)$-coloring, maximal matching, minimal vertex cover, leader election, and consensus under synchronous message passing~\cite{groetschla2025agentsnet}. The framework spans a wide range of models, network topologies, and scales, and highlights clear variation in task difficulty: While some problems like consensus and leader election are solved reliably, others such as coloring and vertex cover remain challenging. Crucially, AgentsNet leaves the design of solution strategies entirely to the LLMs, providing only the problem description and communication channel. This makes it a solid foundation for benchmarking open-ended coordination. Our work builds on this platform, but shifts the focus to controlled analysis by systematically varying the form of symbolic guidance and extending the tasks to harder global variants (e.g.,~3-coloring, maximum matching, minimum vertex cover).

Another recent development is the introduction of \emph{VL-DCOPs}, which extend Distributed Constraint Optimization Problems (DCOPs) to a spectrum of agent types that integrate symbolic and neural components~\cite{mahmud2025distributed}. A DCOP~\cite{modi2005adopt} models coordination as a set of agents, each controlling one or more variables whose value assignments jointly determine the costs of shared constraints. The objective is for the agents to coordinate their assignments so as to minimize the total constraint cost across the network. In the VL-DCOP framework, this coordination spectrum ranges from fully symbolic agents, which execute classical distributed algorithms, to fully neural agents, which rely entirely on LLM-based reasoning. Hybrid variants position LLMs in supporting roles, such as interpreting human instructions, generating constraint models, or resolving uncertainty, while symbolic solvers retain decision-making authority. This notion of hybridity differs from ours: in VL-DCOPs, hybridity arises from combining humans, symbolic solvers, and LLMs, whereas in our work, it emerges from varying the degree of symbolic guidance provided to LLM agents themselves.

More broadly, both VL-DCOPs and our approach touch on a long-standing question in multi-agent systems: How much decision-making power should reside with individual agents versus external controllers? The field of \emph{adjustable autonomy (AA)} emerged to address precisely this issue, proposing that agents dynamically shift control depending on context~\cite{dorais1998adjustable,horvitz1999principles,scerri2002towards}. For example, in a scheduling domain, an agent might act autonomously under routine conditions, but transfer control to a human when uncertainty is high or when coordination failures would be costly~\cite{ferguson1996trading,gunderson1999adjusting}.  

Early approaches often treated autonomy as binary: Either the agent acted independently or it fully delegated to another entity. Recognizing these shortcomings, researchers introduced the notion of \emph{transfer-of-control strategies}~\cite{scerri2001adjustable,scerri2002towards}. These strategies consist of conditional sequences: Agents may attempt an autonomous decision, fall back to external input if needed, and adjust coordination constraints to reduce miscoordination costs. This work demonstrated that \emph{intermediate strategies}, rather than rigid extremes, best preserve both local decision quality and global team coherence. In the next section, we adapt the principle of adjustable autonomy into a taxonomy of guidance strategies for LLM-based agents, treating different forms of symbolic instruction as varying degrees of autonomy.

\section{Symbolic Guidance Taxonomy}\label{sec:SGT}

LLMs can be guided in different ways when solving distributed coordination problems: From being left entirely to their own devices to being constrained by precise symbolic instructions. The \emph{Symbolic Guidance Taxonomy (SGT)} provides a unified framework for describing this design space. It specifies how much of a reference algorithm $\mathcal{A}$ is revealed to agents and in what representational form, thereby defining different points along a spectrum of autonomy. In what follows, we introduce the formal definition of SGT, describe the resulting agent types that instantiate different autonomy levels, and illustrate the taxonomy through a working example.

\subsection{Formal Definition}

The \emph{Symbolic Guidance Taxonomy (SGT)} is a framework for varying which components of a symbolic algorithm are revealed to agents and how they are expressed. AgentsNet~\cite{groetschla2025agentsnet} provided all agents with the same \emph{task description} and \emph{output requirement}, leaving the solution strategy entirely to the LLM agents. Building on this foundation, SGT extends the design space by systematically varying the symbolic structure provided to agents while keeping the task description and output requirement fixed. Specifically, SGT parameterizes the output guidance based on two components: (1) Which parts of an algorithmic description are made available to agents, and (2) The form in which these elements are conveyed.

Formally, let $\mathcal{A}$ denote a symbolic algorithm for solving a given distributed multiagent coordination problem. 
A guidance strategy is defined as a mapping: $G(S, E;\,\mathcal{A}) \;\mapsto\; \mathcal{I}$,
where $S \subseteq \mathcal{A}$ and $E$ together determine the additional guidance $\mathcal{I}$ provided to each agent. In more detail: 
\squishlist
  \item $S$ (\emph{selection}) specifies which parts of the algorithm are supplied, such as a step sequence, a function set, or tie-breaking rules. 
  \item $E$ (\emph{encoding}) specifies how the selected parts are represented, for example as pseudocode, or natural-language descriptions. The encoding is chosen from:
\begin{align*}
  E \in \{& \textsf{main-pseudocode},        \textsf{function-pseudocode}, \\  &\textsf{function-description}, \textsf{full-pseudocode}\},
\end{align*}
  \item $\mathcal{I}$ denotes the resulting \emph{guidance block} received by the agent. 
\squishend

\subsection{Autonomy Spectrum of Agent Types}
\label{sec:agent-types}

This framing connects directly to the principle of \emph{adjustable autonomy}~\cite{scerri2001adjustable,scerri2002towards}. At one extreme, autonomy is maximized: The LLM agent must devise its own procedure without any symbolic guidance. At the other, autonomy is minimized: The LLM agent is asked to execute a fixed protocol with no freedom to deviate. Beyond these bounds lies a purely algorithmic baseline, where autonomy vanishes entirely as the symbolic procedure is executed directly without involving an LLM. Between these extremes, the different guidance strategies are analogous to \emph{transfer-of-control mechanisms} in adjustable autonomy: Different levels of symbolic scaffolding are introduced to reduce miscoordination while still allowing the LLM flexibility to fill in missing details.

\begin{figure*}[t]
  \centering
  \includegraphics[width=1\textwidth]{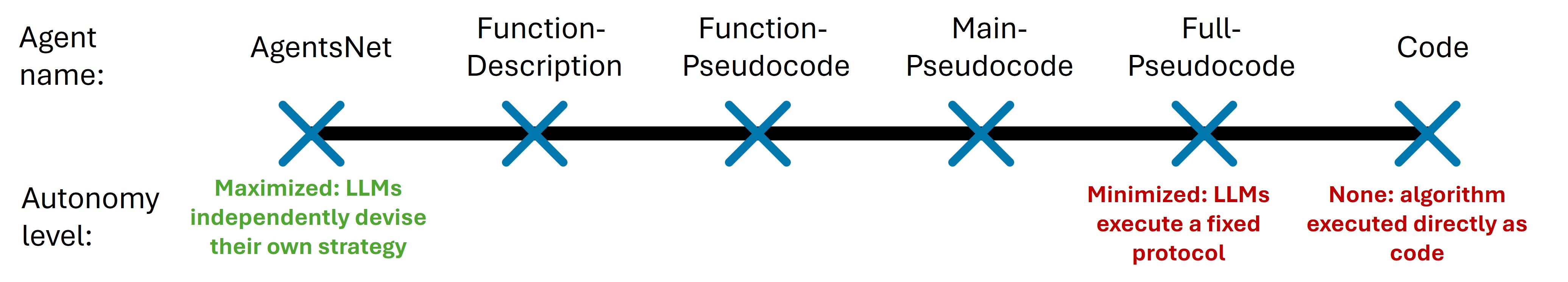}
  \caption{The Symbolic Guidance Taxonomy (SGT) autonomy spectrum, 
  ranging from maximized autonomy (AgentsNet agents) to no autonomy (Code agents).}
  \label{fig:SGT_autonomy}
\end{figure*}

The agent types we define in SGT instantiate this spectrum, ranging from Freeform agents at the high-autonomy end to the Code agent baseline at the no-autonomy end. Each corresponds to a specific $(S,E)$ combination, as illustrated in Figure~\ref{fig:SGT_autonomy}. The five agent types in SGT, corresponding to the five encoding $E$ options, as well as two additional baselines are as follows:

\squishlist
\item \textbf{(Baseline) AgentsNet agent.} No guidance is supplied or, equivalently, $G(\varnothing, \varnothing;\,\mathcal{A}) = \varnothing$.  Agents see only the task description and output requirement, and must devise their own procedure. This corresponds to the baseline setup of AgentsNet~\cite{groetschla2025agentsnet}.

\item \textbf{Function-description agent.} $S$ is the set of all auxiliary helper functions of the algorithm $\mathcal{A}$.\footnote{All functions of the algorithm except for the main function.} Functions are described in natural language, specifying their desired behavior, but not their implementation. Thus, the LLM agents have full flexibility on how to implement them.

\item \textbf{Function-pseudocode agent.} Like Function-description agents, $S$ is the set of all auxiliary helper functions of the algorithm $\mathcal{A}$. However, instead of natural language descriptions, agents receive pseudocode of the functions, effectively suggesting an implementation strategy for those functions. However, note that the LLM agents are not told how those functions should be used to solve the problem.

\item \textbf{Main-pseudocode agent.} Here, $S$ is the main function of the algorithm $\mathcal{A}$ and the agents receive that information in the form of the pseudocode for the function. Note that the main function may call auxiliary helper functions, which are not provided to the LLM agents. Thus, the agents are free to interpret and implement the auxiliary helper functions as they wish.

\item \textbf{Full-pseudocode agent.} $S$ is the complete algorithm $\mathcal{A}$. Agents are provided the complete pseudocode of the algorithm with all steps, function bodies and tie-breaking rules. 

\item \textbf{(Oracle Baseline) Code agent.} As an oracle baseline on the other extreme of this spectrum, we implemented algorithm $\mathcal{A}$ in Python and is executed directly by the agents without any LLM involvement. 

\squishend

\subsection{Illustrative Example}

We now illustrate the SGT using a concrete working example. 
Consider $\mathcal{A}$ as a maximal independent set (MIS) variant for distributed graph coloring (see Section~\ref{sec:background}). At a high level, the MIS-based algorithm proceeds in rounds where agents tentatively propose colors, resolve conflicts with neighbors via tie-breaking rules, and then commit to successful choices. In this example, we focus on the \emph{proposal step} of the algorithm, during which each agent must propose a tentative group assignment. The SGT framework highlights how this same step can take different representational forms depending on the guidance configuration $(S, E)$.

Here, the guidance $\mathcal{I}$ provides minimal scaffolding since it conveys a behavioral intuition without specifying how to operationalize it in code or reasoning.
At the high-autonomy end, the \textbf{Function-description agent} is provided with natural language descriptions of all auxiliary helper functions that support the algorithm. For example:
\emph{``propose\_group: If options remain, choose one at random; otherwise, select the group least frequent among neighbors’ commitments.''} This guidance introduces modular decomposition while preserving linguistic ambiguity.

In contrast, the \textbf{Function-pseudocode agent} receives the same set of auxiliary helper functions but encoded as explicit pseudocode. For instance:
\begin{verbatim}
function propose_group():
    if P(me) not empty:
        tentative <- random choice from P(me)
    else:
        tentative <- least frequent group among neighbors
\end{verbatim}
The pseudocode uses descriptive identifiers and clear structure to minimize ambiguity while retaining readability for LLMs.

Progressing further, the \textbf{Main-pseudocode agent} is given a global outline of the main coordination procedure. Within this outline, the proposal stage appears only as a high-level directive:
\emph{``Step 4: Propose a tentative group (if uncommitted); if none available, select to the least frequent choice among neighbors.''}
This provides the agent with global context but omits detailed implementation.

At the low-autonomy extreme, the \textbf{Full-pseudocode agent} receives the complete algorithm in pseudocode, including both the global structure and all supporting functions. In this setting, the proposal step appears twice: Once in the overall outline (\emph{``Step 4...''}) and again in the detailed helper function (\texttt{propose\_group()}). The resulting specification leaves minimal room for interpretation, effectively guiding the agent toward deterministic execution.

\section{Experimental Design}

Our experiments evaluate how different levels of symbolic guidance affect LLM agents on distributed coordination tasks. We structure the design around five components: The experimental simulator, the graph topologies, the algorithm choices, the LLM models, and the evaluation metrics. 

\paragraph{\textbf{Experimental Simulator:}}
We build on the \emph{AgentsNet} benchmark~\cite{groetschla2025agentsnet},  using their public implementation\footnote{\url{https://github.com/floriangroetschla/AgentsNet}} for message passing and network generation. We provide our own extended implementation,\footnote{\url{https://github.com/YODA-Lab/Symbolic-Guidance-Taxonomy}} which adds harder global variants (3-coloring, maximum matching, minimum vertex cover) and systematically varies symbolic guidance through the Symbolic Guidance Taxonomy (SGT).

\paragraph{\textbf{Graph Topologies:}} Agents operate on three graph families generated with the original benchmark code using NetworkX:\footnote{\url{https://networkx.org/}}
\squishlist
    \item \emph{Smallworld}~\cite{watts1998collective}: Strong local clustering with a few random shortcuts.
    \item \emph{Scalefree}~\cite{barabasi1999emergence}: Hub-dominated networks with power-law degree distributions.
    \item \emph{Delaunay}~\cite{lee1980two}: Planar graphs emphasizing spatial locality. 
\squishend

For each topology, we generate $5$ graphs of size $n=16$, yielding $15$ networks. All experiments run for $2D{+}1$ synchronous message-passing rounds (where $D$ is the graph diameter), ensuring each agent can indirectly reach all other agents. We evaluate all agent types from SGT, as described in Section~\ref{sec:agent-types}. Each experiment uses the same number of message-passing rounds, isolating the effect of symbolic guidance.

\paragraph{\textbf{Algorithm Choices:}} The algorithm $\mathcal{A}$, whose components are provided as guidance to the LLM agents, depend on the variant of the problem domain. For local variants (i.e.,~($\Delta$ + 1)-coloring, maximal matching, and minimal vertex cover), we use the \emph{Maximal Independent Set (MIS) heuristic}. For the global variants (i.e.,~3-coloring, maximum matching, and minimum vertex cover), we use the MIS heuristic as well as the \emph{Distributed DFS algorithm}. See Section~\ref{sec:algorithms} for more information of the two algorithms. 


\paragraph{\textbf{LLM Models:}} We evaluate three representative LLM models, covering both free open-source options and a more expensive commercial system:
\squishlist
    \item \emph{Llama-3.1-8B Instruct} (Meta), served locally via Ollama. 
    \item \emph{Qwen-2.5-7B Instruct} (Alibaba), also via Ollama. 
    \item \emph{Gemini-2.5 Flash} (Google DeepMind), accessed via OpenRouter. Identified by \emph{AgentsNet} as Pareto-optimal in accuracy-cost trade-offs~\cite{groetschla2025agentsnet}.
\squishend

\paragraph{\textbf{Evaluation Metrics:}} 
We adopt a unified error metric, the \emph{Sum of Squared Error (SSE)}, which captures deviations from correctness or optimality. We define them below for each of our three domains:

\squishlist
\item \emph{Graph coloring}: For a solution $x$, an edge is \emph{valid} if its vertices have different colors in the solution, and \emph{invalid} otherwise. Let $V(x)$ and $I(x)$ denote the sets of valid and invalid edges in the solution, respectively, and $|OPT(G)|$ denote the maximum number of edges that can be properly colored in the input graph $G$. Using these definitions, the SSE is defined as follows for the two local and global variants of this problem:
\squishlist
    \item ($\Delta$ + 1)-Coloring: $\mathrm{SSE}_{\mathrm{coloring}}^{\mathrm{local}}(x) = |I(x)|^2$.
    \item 3-Coloring: 
    $\mathrm{SSE}_{\mathrm{coloring}}^{\mathrm{global}}(x) = \bigl(|OPT(G)| - |V(x)|\bigr)^2$.
\squishend

\item \emph{Matching}: For a solution $x$, let $M(x)=\{\{u,v\}\in E \mid x(u)=v \ \land\ x(v)=u\}$ define the set of matches in the solution.
An edge is \emph{valid} if it belongs to $M(x)$ or if at least one of its vertices is already matched via another edge in $M(x)$ (so the edge is indirectly covered). Otherwise it is \emph{invalid}. Let $I(x)$ denotes the set of invalid edges in the solution and $|OPT(G)|$ denote the number of a matches in an optimal solution of the problem.
\squishlist
    \item Maximal Matching: $\mathrm{SSE}_{\mathrm{matching}}^{\mathrm{local}}(x) = |I(x)|^2$.
    \item Maximum Matching: $\mathrm{SSE}_{\mathrm{matching}}^{\mathrm{global}}(x) = \bigl(|OPT(G)| - |M(x)| \bigr)^2$. 
\squishend

\item \emph{Vertex cover}: 
For a solution $x$, a vertex is \emph{valid} if it is in the selected set and is required to cover at least one uncovered edge (i.e., none of its neighbors are already in the cover), and \emph{invalid} otherwise. 
Let $V(x)$ and $I(x)$ denote the sets of valid and invalid vertices in the solution, respectively, and $|OPT(G)|$ denote the minimum number of vertices needed to optimally cover all edges in the problem.
\squishlist
    \item Minimal Vertex Cover:  
    $\mathrm{SSE}_{\mathrm{cover}}^{\mathrm{local}}(x) = |I(x)|^2$.
    \item Minimum Vertex Cover: 
    $\mathrm{SSE}_{\mathrm{cover}}^{\mathrm{global}}(x) = \bigl(|OPT(G)| - |V(x)|\bigr)^2 + |I(x)|^2$.
\squishend
\squishend

\section{Experimental Results}

\begin{figure*}[t]
  \centering
  \includegraphics[width=\textwidth]{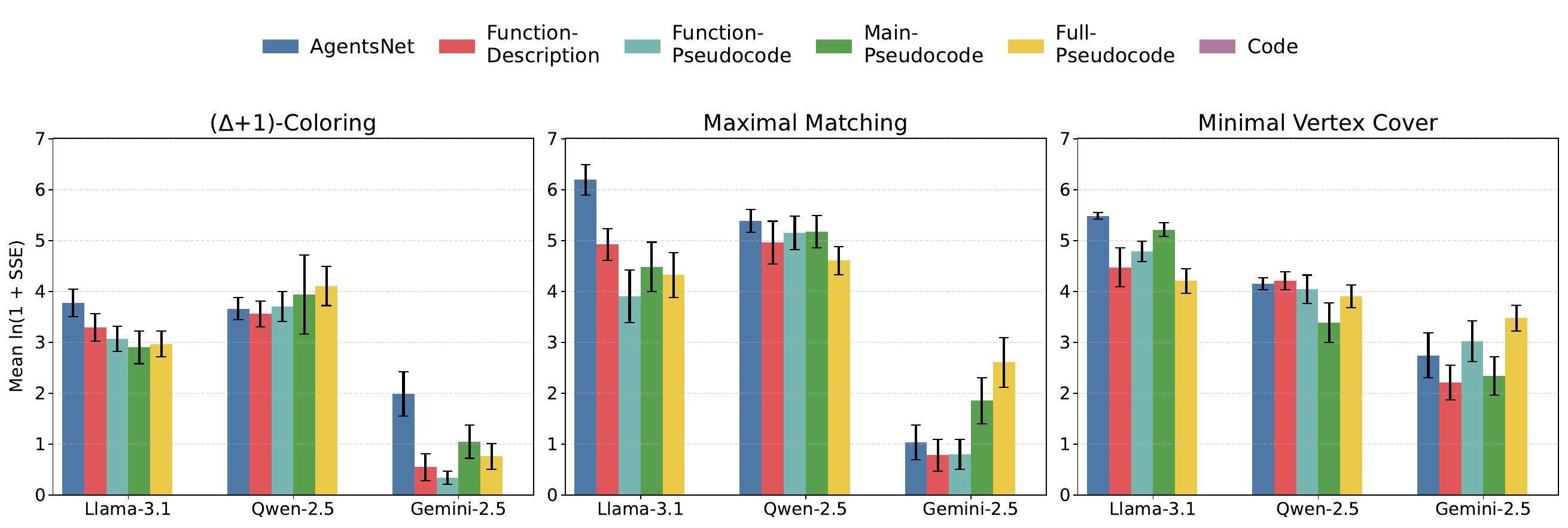}
  \caption{Local coordination domains with MIS as algorithm $\mathcal{A}$. Showing mean $\ln(1+\mathrm{SSE})$ by model. Bar colors denote agent types sorted by autonomy level; error bars show SEM.}
  \label{fig:easy-error}
\end{figure*}

\begin{figure}[t]
  \centering
  \includegraphics[width=0.5\textwidth]{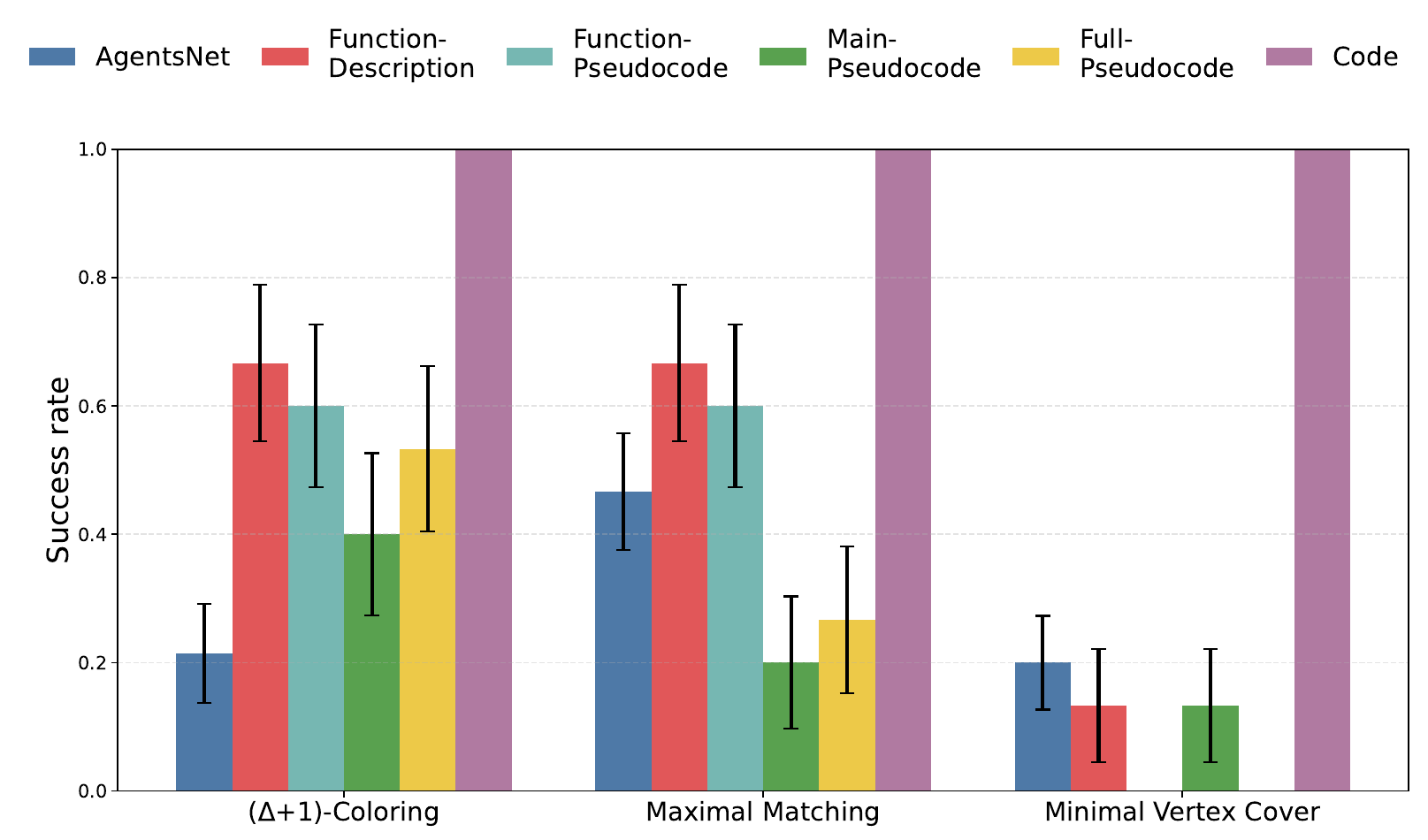}
 \caption{Local coordination domains with MIS as algorithm $\mathcal{A}$. Showing the success rate of feasible solutions for Gemini. Bar colors denote agent types sorted by autonomy level; error bars show SEM.}
  \label{fig:easy-success}
\end{figure}

\begin{figure*}[t]
  \centering
  \includegraphics[width=\textwidth]{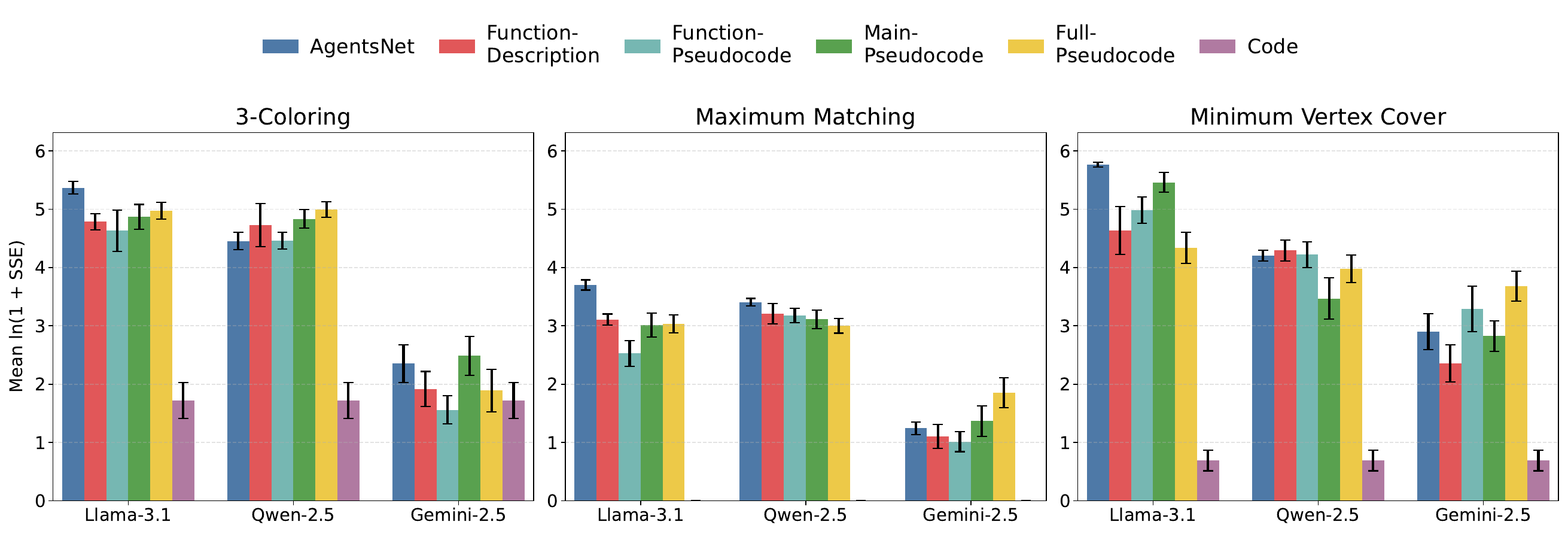}
  \caption{Global coordination domains with MIS as algorithm $\mathcal{A}$. Showing mean $\ln(1+\mathrm{SSE})$ by model. Bar colors denote agent types sorted by autonomy level; error bars show SEM.}
  \label{fig:hard-MIS}
\end{figure*}

\begin{figure*}[t]
  \centering
  \includegraphics[width=\textwidth]{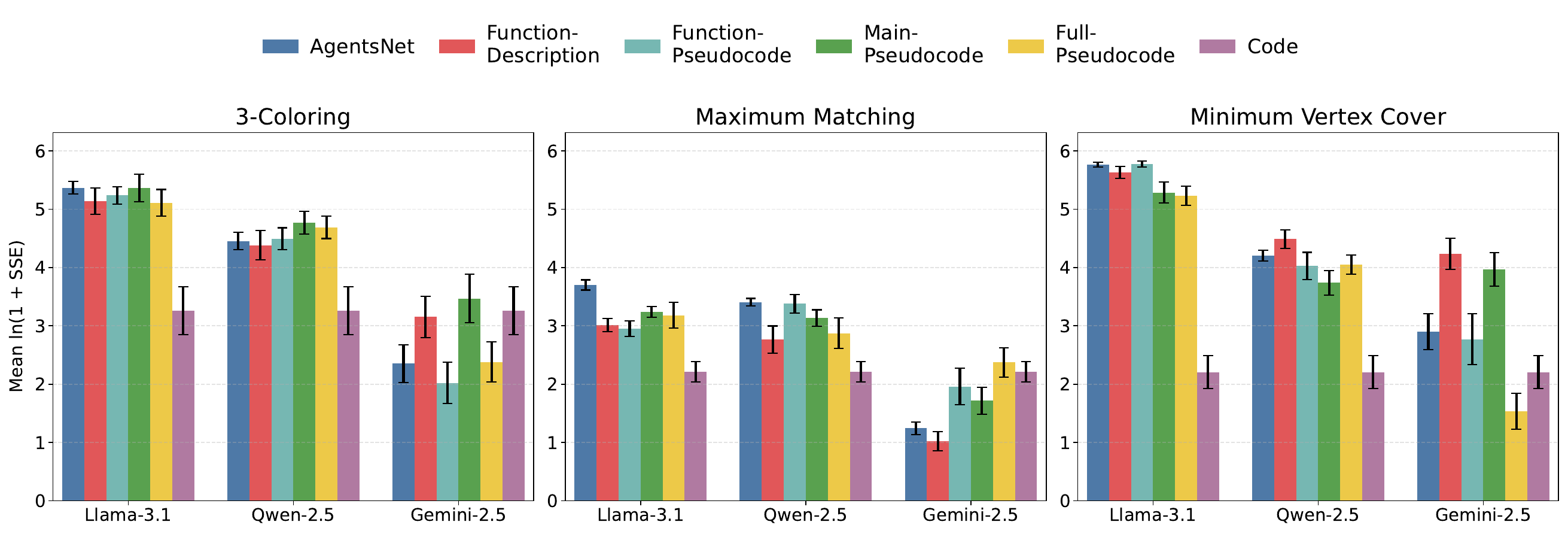}
  \caption{Global coordination domains with Distributed DFS as algorithm $\mathcal{A}$. Showing mean $\ln(1+\mathrm{SSE})$ by model. Bar colors denote agent types sorted by autonomy level; error bars show SEM.}
  \label{fig:HARD- COMPLETE SEARCH}
\end{figure*}

We now describe our experimental results on the performance of the different agent types across the SGT autonomy spectrum (see Section~\ref{sec:agent-types}) and different domain variants (see Section~\ref{sec:domains}).

\subsection{Local Coordination Problems}

We first focus on the local variants of the three domains, where MIS is the algorithm $\mathcal{A}$ whose components are provided as guidance to the LLM agents. 
Figure~\ref{fig:easy-error} shows the average error of all the five SGT agent types as well as the two AgentsNet and Code baselines with all three LLM models across the three domains. We make the following observations:
\squishlist
\item In general, the Gemini agents perform better than their Llama and Qwen counterparts, highlighting that Gemini is a better LLM for distributed graph coordination problems. This observation is consistent with similar findings in the literature for other problem types, such as open-ended natural-language dialogue (e.g., explaining code, summarizing text, or answering creative prompts) and multimodal academic reasoning (e.g., college-level question answering that integrates text with images such as diagrams, charts, and tables across multiple disciplines)~\cite{chiang2024chatbot,yue2024mmmu}.

\item On whether symbolic guidance improves on the AgentsNet baseline, there are mixed results. Specifically,
\squishlist
\item When the agents use Llama models, all forms of guidance help with little statistical significance differences between the different agent types. However, they are all statistically significantly better than AgentsNet.
\item When the agents use Qwen models, there is mostly no statistical differences between AgentsNet and all the different SGT agent types.
\item When the agents use Gemini models, all forms of guidance help in ($\Delta$ + 1)-coloring problems. \ben{In the maximal matching and minimal vertex cover problems, the function-description scaffold yields a slight additional improvement.} Interestingly, some agent types perform worse than AgentsNet in some settings (e.g.,~Main- and full-pseudocode agents in maximal matching problems), highlighting that the guidance provided can some times do more harm than good. 
\squishend
\squishend

Figure~\ref{fig:easy-success} shows the success rates\footnote{Defined as the fraction of instances that were successfully solved.} of all the agent types with the Gemini model only. We omitted results for the Llama and Qwen as all variants except for the oracle Code baseline failed to solve all instances in these domains. We make the following observations:
\squishlist
\item As expected, the oracle Code agent achieves perfect success rates across all three domains. 
\item For ($\Delta$ + 1)-coloring problems, consistent with results from Figure~\ref{fig:easy-error}, all forms of guidance improve the success rate of AgentsNet with function-based guidance (i.e.,~function-pseudocode and function-description agents) performing the best. 
\item For maximal matching problem, function-based guidance also perform the best. 
\item \ben{Finally, for minimal vertex cover problems, guidance did not appear to produce meaningful differences in the number of feasible instances.}
\squishend

\subsection{Global Coordination Problems}

We now present results on the global variants of the three domains, where either MIS or Distributed DFS is the algorithm $\mathcal{A}$ whose components are provided as guidance to the LLM agents. 
For these experiments, we constraint the oracle Code agent to use the same number of communication rounds as the LLM agents to allow for fair comparisons between the different agent types.

Figure~\ref{fig:hard-MIS} shows the results when the algorithm employed is MIS. In general, the trends here are mostly consistent with the trends from the local variants shown in Figure~\ref{fig:easy-error}: 
\squishlist
\item Gemini agents perform better than Llama and Qwen agents across all three problem domains.
\item Among Llama agents, in general, all forms of guidance improve upon AgentsNet with statistical significance. 
\item Among Qwen agents, in general, there is no statistical difference between AgntsNet and the SGT agent types. 
\item Among Gemini agents, Function-pseudocode agents perform best in 3-coloring and maximum matching, while Function-description agents perform best in minimum vertex cover. 
\squishend
The key difference with the results from the local variants is that the oracle Code agent now has a non-zero error. The main reason is the following: While MIS is guaranteed to find a feasible solution in the simpler local variant problems, it is not guaranteed to find optimal solutions in these problems that require global coordination. 

Figure~\ref{fig:HARD- COMPLETE SEARCH} shows the results when the algorithm employed is Distributed DFS. Most of the trends from earlier still apply here as well. However, there are several key differences:
\squishlist
\item Some of the Gemini agent types perform better than the oracle Code agent! While one expects that oracle Code agents find optimal solutions because they are running Distributed DFS, which has optimality guarantees, that is under the assumption that it has enough runtime to run till convergence. As we limit the number of communication rounds to be the same as that of the LLM agents, this guarantee no longer applies, and the reported results are not necessarily optimal.\footnote{While the same constraint is imposed on Code agents when it is running MIS, that limitation did not prevent it from finding the best solutions across the different agent types.}  
\item When using Gemini models, none of the guidance forms help in 3-coloring problems. However function-description guidance helped in maximum matching and full-pseudocode guidance helped in minimum vertex cover problems. 
\squishend

\ben{As Gemini consistently achieves the lowest errors across all domain variants and algorithms (Figures~\ref{fig:easy-error}–\ref{fig:HARD- COMPLETE SEARCH}) among the three LLM models, we focus our detailed quantitative analysis on this model. Table~\ref{tab:gemini-summary} consolidates these results across all tasks and algorithms, showing that, in most cases, \emph{the most effective agents are those with function-based guidance} (i.e.,~function description and function pseudocode) \emph{exhibiting intermediate autonomy}.}

\subsection{Discussion}

Across all domains, the results reveal that both the form of symbolic guidance and the underlying algorithm jointly shape agents' ability to reason and coordinate effectively. 

Among the tested models, Gemini consistently demonstrates the strongest capacity to leverage symbolic structure, achieving lower errors and more stable performance across guidance levels. This trend suggests that more capable models are better able to internalize and operationalize symbolic scaffolds, translating external structure into coherent multiagent reasoning.

Notably, function description agents often achieve the best or near-best results across domains. This suggests that LLMs perform best when symbolic guidance captures algorithmic intent at a conceptual level rather than prescribing every procedural detail. In contrast, both unguided AgentsNet agents and fully specified pseudocode agents tend to underperform, indicating that excessive autonomy or over-specification can each disrupt coherent coordination.

For local agent-based MIS heuristics, lightweight scaffolds (e.g.,~function pseudocode) further illustrate this pattern, providing enough structure to support coordination while preserving reasoning flexibility. In contrast, overly detailed guidance (e.g., full pseudocode specifications) often leads to degraded performance, implying that symbolic scaffolds are most effective when they complement rather than constrain the model's reasoning process.

For the global search-based Distributed DFS algorithm, results are more mixed: Performance differences across SGT agent types are often not statistically significant. In some cases, detailed guidance helps (e.g., full-pseudocode agents with Gemini on minimum vertex cover problems), while in others, it hinders performance (e.g., function-description and main-pseudocode agents on the same task). Overall, these findings suggest that \emph{LLM agents benefit most from symbolic scaffolds that align with simpler, local heuristics}, whose algorithmic structure can be more readily absorbed and integrated by language models, unlike complex global search procedures that impose rigid control flows and exceed the model's representational strengths.

\begin{table}[t]
\centering
\small
\setlength{\tabcolsep}{4pt}
\renewcommand{\arraystretch}{1.05}
\begin{tabular}{lccc}
\toprule
\multicolumn{4}{c}{\emph{Local (MIS)}}\\
\cmidrule(lr){1-4}
& \textbf{($\Delta$+1)-} & \textbf{Maximal} & \textbf{Minimal} \\[-1pt]
& \textbf{Coloring} & \textbf{Matching} & \textbf{Vertex Cover} \\
AgentsNet        & 1.99 $\pm$ 0.43 & 1.03 $\pm$ 0.34 & 2.74 $\pm$ 0.44 \\
Function-Description       & 0.55 $\pm$ 0.26 & \textbf{0.78 $\pm$ 0.31} & \textbf{2.21 $\pm$ 0.34} \\
Function-Pseudocode       & \textbf{0.34 $\pm$ 0.12} & 0.80 $\pm$ 0.30 & 3.03 $\pm$ 0.40 \\
Main-Pseudocode           & 1.05 $\pm$ 0.33 & 1.85 $\pm$ 0.46 & 2.34 $\pm$ 0.38 \\
Full-Pseudocode           & 0.76 $\pm$ 0.25 & 2.61 $\pm$ 0.49 & 3.48 $\pm$ 0.25 \\
\midrule
\multicolumn{4}{c}{\emph{Global (MIS)}}\\
\cmidrule(lr){1-4}
& \textbf{3-Coloring} & \textbf{Maximum} & \textbf{Minimum} \\[-1pt]
&  & \textbf{Matching} & \textbf{Vertex Cover} \\
AgentsNet        & 2.35 $\pm$ 0.33 & 1.24 $\pm$ 0.11 & 2.90 $\pm$ 0.31 \\
Function-Description       & 1.92 $\pm$ 0.30 & 1.11 $\pm$ 0.20 & \textbf{2.35 $\pm$ 0.32} \\
Function-Pseudocode       & \textbf{1.56 $\pm$ 0.24} & \textbf{1.01 $\pm$ 0.17} & 3.29 $\pm$ 0.39 \\
Main-Pseudocode           & 2.49 $\pm$ 0.33 & 1.37 $\pm$ 0.26 & 2.83 $\pm$ 0.26 \\
Full-Pseudocode           & 1.89 $\pm$ 0.36 & 1.85 $\pm$ 0.25 & 3.68 $\pm$ 0.26 \\
\midrule
\multicolumn{4}{c}{\emph{Global (Distributed DFS)}}\\
\cmidrule(lr){1-4}
& \textbf{3-Coloring} & \textbf{Maximum} & \textbf{Minimum} \\[-1pt]
&  & \textbf{Matching} & \textbf{Vertex Cover} \\
AgentsNet        & 2.35 $\pm$ 0.33 & 1.24 $\pm$ 0.11 & 2.90 $\pm$ 0.31 \\
Function-Description       & 3.15 $\pm$ 0.36 & \textbf{1.02 $\pm$ 0.16} & 4.23 $\pm$ 0.27 \\
Function-Pseudocode       & \textbf{2.02 $\pm$ 0.36} & 1.96 $\pm$ 0.32 & 2.77 $\pm$ 0.44 \\
Main-Pseudocode           & 3.47 $\pm$ 0.42 & 1.72 $\pm$ 0.23 & 3.97 $\pm$ 0.29 \\
Full-Pseudocode           & 2.38 $\pm$ 0.34 & 2.37 $\pm$ 0.25 & \textbf{1.54 $\pm$ 0.31} \\
\bottomrule
\end{tabular}
\vspace{2mm}
\caption{Gemini-2.5: mean $\ln(1+\mathrm{SSE}) \pm \mathrm{SEM}$ across different experimental settings (best per column in \textbf{bold}). Each section header indicates the domain variant and, in parentheses, the algorithm used.}

\label{tab:gemini-summary}
\end{table}

\section{Conclusions}

This paper introduced the Symbolic Guidance Taxonomy (SGT) -- a principled framework for analyzing how symbolic structure shapes intelligence and coordination in LLM-based multiagent systems. SGT conceptualizes agent behavior as a continuum of autonomy, identifying symbolic guidance as a controllable mechanism that modulates this autonomy in distributed multiagent coordination settings.

Empirically, we evaluated the SGT framework across multiple coordination domains (graph coloring, matching, and vertex cover). We examined six LLM agent types positioned along the SGT autonomy spectrum and contrasted two algorithmic paradigms: Local heuristic-based coordination and global exhaustive distributed search. Together, these experiments provide a systematic view of how symbolic representation and autonomy interact to influence coordination quality.

Our results show that coordination performance does not improve monotonically with either increased guidance or greater autonomy; rather, it depends on how the two are balanced. \emph{The strongest gains occur at intermediate autonomy levels, where symbolic structure provides a useful scaffold for reasoning while maintaining flexibility}. These benefits, however, arise primarily when the guidance derives from heuristic algorithms, whose simpler and more interpretable structures align well with the reasoning capabilities of LLM agents. In contrast, guidance based on global search algorithms typically yields little to no improvement, likely because such information is overly detailed and complex, making it difficult for LLM agents to distill and effectively exploit during coordination.

These findings demonstrate that symbolic guidance serves as a principled lever for regulating autonomy in LLM agents. Rather than treating autonomy as fixed, SGT enables it to be tuned through the form and representation of symbolic information, allowing system designers to balance reliability and adaptability in multiagent coordination. This connection bridges classical distributed algorithm design with emergent language-based reasoning, showing that formal symbolic control and generative flexibility can coexist within a unified multiagent framework. Future work will extend SGT to mixed-autonomy teams, where agents within the same graph may operate at different levels of autonomy depending on their role or local context. We aim to develop adaptive guidance mechanisms that allow agents to adjust their autonomy dynamically as tasks unfold and coordination demands shift.

\subsubsection*{Acknowledgements.}

This work was partially supported by the Office of Naval Research (N00014-24-1-2663).

\bibliographystyle{splncs04}
\bibliography{refs}
\end{document}